\documentclass[10pt,twocolumn,letterpaper]{article}
\usepackage[accsupp]{axessibility}
\usepackage{cvpr}              % To produce the CAMERA-READY version
\usepackage{graphicx}
\usepackage{amsmath}
\usepackage{amssymb}
\usepackage{booktabs}

\usepackage[pagebackref,breaklinks,colorlinks]{hyperref}

\usepackage[capitalize]{cleveref}
\crefname{section}{Sec.}{Secs.}
\Crefname{section}{Section}{Sections}
\Crefname{table}{Table}{Tables}
\crefname{table}{Tab.}{Tabs.}

\def\cvprPaperID{4182} % *** Enter the CVPR Paper ID here
\def\confName{CVPR}
\def\confYear{2023}

\usepackage{graphics} % for pdf, bitmapped graphics files
\usepackage{epsfig} % for postscript graphics files
\usepackage{mathptmx} % assumes new font selection scheme installed
\usepackage{times} % assumes new font selection scheme installed
\usepackage{amsmath} % assumes amsmath package installed
\usepackage{amssymb}  % assumes amsmath package installed
\usepackage{color}
\usepackage{multirow}
\usepackage{comment}
\usepackage{physics}
\usepackage{tabu}
\usepackage{cite}

\newcommand\blfootnote[1]{%
  \begingroup
  \renewcommand\thefootnote{}\footnote{#1}%
  \addtocounter{footnote}{-1}%
  \endgroup
}

\newcommand{\tabincell}[2]{\begin{tabular}{@{}#1@{}}#2\end{tabular}}

\begin{document}

\title{
Hierarchical Temporal Transformer for 3D Hand Pose Estimation and Action Recognition from Egocentric RGB Videos}

\author{Yilin Wen$^{1}$ Hao Pan$^{2}$ Lei Yang$^{3,1}$ Jia Pan$^{1}$ Taku Komura$^{1}$ Wenping Wang$^{4}$\\
$^{1}$The University of Hong Kong $^{2}$ Microsoft Research Asia $^{3}$ TransGP $^{4}$ Texas A{\&}M University\\
{\tt\small \{ylwen,jpan,taku\}@cs.hku.hk haopan@microsoft.com l.yang@transgp.hk  wenping@tamu.edu}}
\maketitle

%%%%%%%%% ABSTRACT
\begin{abstract}
Understanding dynamic hand motions and actions from egocentric RGB videos is a fundamental yet challenging task due to self-occlusion and ambiguity. 
To address occlusion and ambiguity, we develop a transformer-based framework to exploit temporal information for robust estimation. 
Noticing the different temporal granularity of and the semantic correlation between hand pose estimation and action recognition, we build a network hierarchy with two cascaded transformer encoders, where the first one exploits the short-term temporal cue for hand pose estimation, and the latter aggregates per-frame pose and object information over a longer time span to recognize the action. 
Our approach achieves competitive results on two first-person hand action benchmarks, namely FPHA and H2O. 
Extensive ablation studies verify our design choices. 

\blfootnote{Work is partially done during the internship of Y. Wen with Microsoft Research Asia. Code and data are available at \url{https://github.com/fylwen/HTT}.}
\end{abstract}

%%%%%%%%% BODY TEXT
\section{Introduction}
\begin{figure}[!tb]
\centering
\includegraphics[width=0.99\linewidth]{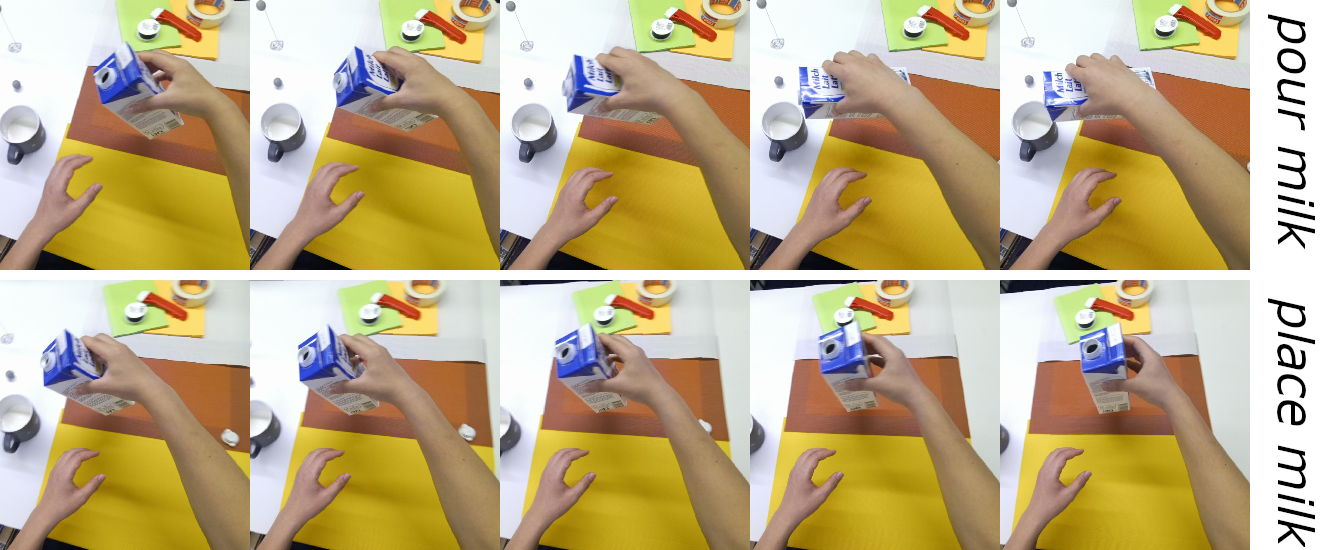}
\caption{Image sequences for \textit{pour milk} and \textit{place milk} under egocentric view from H2O~\cite{kwon2021h2o}, with frequently occluded hand joints and ambiguous action type judged by individual frames. Using temporal information can benefit both tasks of 3D hand pose estimation and action recognition.}\label{fig:teasor}
\end{figure}

Perceiving dynamic interacting human hands is fundamental in fields such as human-robot collaboration, imitation learning, and VR/AR applications. 
Viewing through the egocentric RGB video is especially challenging, as there are frequent self-occlusions between hands and objects, as well as severe ambiguity of action types judged from individual frames (\textit{e.g.} see Fig.~\ref{fig:teasor} where the actions of \textit{pour milk} and \textit{place milk} can only be discerned at complete sequences).

Recent years have witnessed tremendous improvement in 3D hand pose estimation and action recognition. 
While many works focus on only one of these  tasks~\cite{zimmermann2017learning,iqbal2018hand,fan2020adaptive,moon2020interhand2,hasson2020leveraging, feichtenhofer2016convolutional,feichtenhofer2019slowfast,yan2018spatial}, unified frameworks~\cite{tekin2019h+,kwon2021h2o,yang2020collaborative} have also been proposed to address both tasks simultaneously, based on the critical observation that the temporal context of hand poses helps resolve action ambiguity, implemented via models like LSTM, graph convolutional network or temporal convolutional network.
However, we note that temporal information can also benefit hand pose estimation: while interacting hands are usually under partial occlusion and truncation, especially in the egocentric view, they can be inferred more reliably from neighboring frames with different views by temporal motion continuity.
Indeed, this idea has not been fully utilized yet among the existing works~\cite{tekin2019h+,kwon2021h2o,yang2020collaborative}: \textit{e.g.}, \cite{tekin2019h+,kwon2021h2o} perform hand pose estimation at each frame, leaving the temporal dimension unexplored, and~\cite{yang2020collaborative} jointly refines action and hand pose through hand-crafted multiple-order motion features and a complex iterative scheme.

We build a simple end-to-end trainable framework to exploit the temporal dimension and achieve effective hand pose estimation and action recognition with a single feed-forward pass. 
To exploit the relationship among frames, we adopt the transformer architecture~\cite{vaswani2017attention} which has demonstrated superior performance in sequence modeling.
However, action and pose have different temporal granularity: while the action is related to longer time spans lasting for several seconds, the hand pose depicts instantaneous motions.
Correspondingly, we use two transformer encoders with different window sizes to respectively leverage the short-term and long-term temporal cues for the per-frame pose estimation and the action recognition of a whole sequence.
Moreover, we notice that the action has a higher semantic level, which is usually defined in the form of ``\textit{verb + noun}"~\cite{garcia2018first,kwon2021h2o}, where \textit{verb} can be derived from the hand motion and the \textit{noun} is the object being manipulated. We thus follow this pattern to build a hierarchy by cascading the pose and action blocks, where the pose block outputs the per-frame hand pose and object label, which are then aggregated by the action block for action recognition.

We evaluate our approach on FPHA\cite{garcia2018first} and H2O\cite{kwon2021h2o}, and achieve state-of-the-art performances for 3D hand pose estimation and action recognition from egocentric RGB videos. Our contribution is summarized as follows:
\begin{itemize}
    \itemsep0em
    \item We propose a simple but efficient end-to-end trainable framework to leverage the temporal information for 3D hand pose estimation and action recognition from egocentric RGB videos. 
    \item We build a hierarchical temporal transformer with two cascaded blocks, to leverage different time spans for pose and action estimation, and model their semantic correlation by deriving the high-level action from the low-level hand motion and manipulated object label.
    \item We show state-of-the-art performance on two public datasets FPHA\cite{garcia2018first} and H2O\cite{kwon2021h2o}.
\end{itemize}

\section{Related Work}
\noindent\textbf{3D hand pose estimation from monocular RGB image/video } 
Massive literature focuses only on the task of 3D pose estimation for hand (or two hands) recorded in the monocular RGB input, to output the 3D skeleton~\cite{zimmermann2017learning,iqbal2018hand,spurr2020weakly,moon2020interhand2,kim2021end,meng20223d,mueller2018ganerated,cai2019exploiting,fan2020adaptive}, or reconstruct also the hand mesh~\cite{boukhayma20193d,zhang2019end,park2022handoccnet,zhang2021interacting,hampali2022keypoint,wang2020rgb2hands,han2020megatrack,ge20193d,choi2020pose2mesh,chen2021camera,li2022interacting,lin2021end,lin2021mesh}.
While most of these articles exploit only the spatial dimensions, ~\cite{mueller2018ganerated,cai2019exploiting,fan2020adaptive,han2020megatrack,wang2020rgb2hands} work on the video-based setting and leverage also temporal consistency for improved robustness, using models such as LSTM in~\cite{fan2020adaptive} or graph convolutional network in~\cite{cai2019exploiting}.

Same as ours,~\cite{tekin2019h+,kwon2021h2o,yang2020collaborative} simultaneously tackle tasks of hand pose estimation and dynamic action recognition from video. However, H+O~\cite{tekin2019h+} and H2O~\cite{kwon2021h2o} perform image-based hand pose estimation by outputting the 3D joint position based on a 3D grid representation, leaving the temporal consistency unexplored.
Collaborative~\cite{yang2020collaborative} initializes the estimation by deriving per-joint 2D heatmaps and depth maps, and iteratively refines the hand pose with temporal action prior. 
In comparison, we use hierarchical transformers to exploit short-term temporal cues for hand pose estimation and long-term temporal cues for action recognition, with all computations done in a simple and efficient feed-forward pass.

\noindent\textbf{Action recognition from monocular RGB video }
In addition to hand pose, interests grow in recognizing semantic hand actions where objects are manipulated \cite{garcia2018first,kwon2021h2o}. 
To recognize the performed hand action (or more general human actions), temporal information is usually necessary. 
For example,~\cite{simonyan2014two,feichtenhofer2016convolutional,carreira2017quo} build on a two-stream convolutional network to exploit the spatial-temporal features from the input frames, while~\cite{feichtenhofer2019slowfast,feichtenhofer2020x3d} leverage on different frame rates for improved flexibility and efficiency. 
However, they neither stress the pose estimation nor utilize the semantic relationship between pose and action. 
On the other hand, follow-up research has shown the benefits of explicitly modeling the pose-action relationship and deriving the action type from motion features. For example, ~\cite{li2015delving,ma2016going,singh2016first} and ~\cite{liu2020forecasting,dessalene2021forecasting} leverage the 2D hand masks or locations to better understand the hand-object interaction, which further facilitates action recognition or anticipation. Another group of works refers to the 3D skeleton features for action recognition, based on temporal models such as temporal convolutional network in~\cite{soo2017interpretable,ke2017new, luvizon20182d,yang2020collaborative}, 
LSTM in~\cite{liu2016spatio,liu2017global,tekin2019h+},  
and graph convolutional network in~\cite{yan2018spatial,shi2019two,kwon2021h2o}.

Among works that simultaneously process both 3D hand pose estimation and action recognition from the given input video, Collaborative~\cite{yang2020collaborative} aggregates the slow-fast features for multiple orders in the temporal dimension and multi-scale relations on the topology of hand skeleton, and outputs the action category with the temporal convolutional network. 
H+O~\cite{tekin2019h+} first outputs the per-frame pose for the hand and manipulated object (if applicable), as well as the verb category and noun category for action, and integrate the per-frame estimations across the video based on an LSTM framework for improved action recognition. H2O~\cite{kwon2021h2o} improves H+O~\cite{tekin2019h+} with a topology-aware graph convolutional network (TA-GCN) to simultaneously model both the inter-frame and inner-frame relationship for hand object interaction. 
We follow the semantic relationship to derive action from hand motion and object label, but use a more powerful hierarchical transformer framework to adaptively attend to frames simultaneously, for robust pose and action estimation (see Fig.~\ref{fig:attn_action}). 

\noindent\textbf{Transformers } Transformers~\cite{vaswani2017attention} prevail in natural language processing and computer vision. 
With multi-head self-attention, this framework effectively models relationships among different tokens of an input sequence. 
For hand pose estimation, the transformer framework has been adopted mainly to capture interactions of image features and key joints in the spatial domain~\cite{huang2020hand,lin2021end,lin2021mesh,park2022handoccnet,huang2020hot,meng20223d,li2022interacting}. Meanwhile, ~\cite{yang2021beyond, arnab2021vivit} adopt clip-level transformers with video-level fusion, to leverage the coherence of clips and the whole sequence for robust action recognition.

We adopt the transformer framework to simultaneously achieve tasks at two levels (pose estimation and action recognition), by respecting their different temporal granularities and modeling their semantic dependency.
Moreover, inspired by~\cite{liu2021swin} that builds a hierarchy of shifted windows to exploit multi-scale features within the image plane, we extend the hierarchical shifting window strategy in the context of multitasking, to learn temporal models of different granularities and model the dependant semantic relationship with efficient computation.

\section{Methodology}
\begin{figure*}[!tb]
\centering
\includegraphics[width=0.85\linewidth]{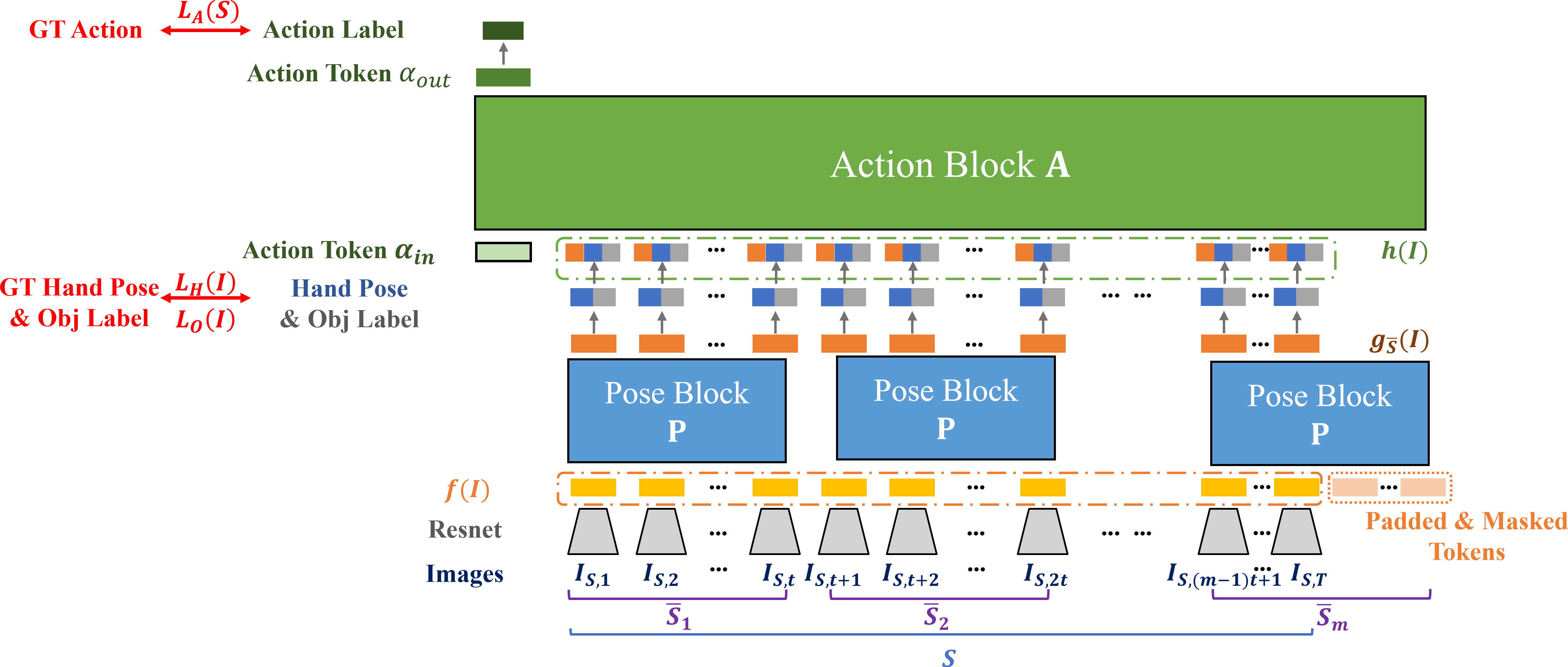}
\vspace{-2mm}
\caption{Overview of our framework. Given input video $S$, we first feed each image to a ResNet feature extractor, and then leverage short-term temporal cue via $\vb{P}$ applied to shifted windowed frames, to estimate per-frame 3D hand pose and object label. We finally aggregate the long-term temporal cue with $\vb{A}$, to predict the performed action label for $S$ from the hand motion and manipulated object label. We supervise the learning with GT labels.}\label{fig:network}
\end{figure*}

Given a first-person RGB video recording the dynamic hand (or two-hands) performing an action, we estimate the per-frame 3D hand pose and the action category from a given taxonomy \cite{garcia2018first,tekin2019h+,yang2020collaborative,kwon2021h2o}. 

Our network is visualized in Fig.~\ref{fig:network}.
To encode the spatial information of each frame, we first feed each frame to a ResNet-18~\cite{he2016deep} and obtain its feature vector from the last layer of the ResNet before softmax. 
To leverage temporal information, the sequence of frame features is passed to our hierarchical temporal transformer (Sec.~\ref{sec:htt}), which is a cascaded framework that uses different time spans for per-frame 3D hand pose estimation (Sec.~\ref{sec:pose}) and action recognition (Sec.~\ref{sec:action}).
To implement the different time spans efficiently, we adopt the shifting window strategy (see Fig.~\ref{fig:sw}) to split the video into sub-sequences for pose estimation and action recognition. 

\subsection{Hierarchical Temporal Transformer}\label{sec:htt}
Our core design is the hierarchical temporal transformer ($\vb{HTT}$, see Fig.~\ref{fig:network}), which exploits temporal cues from the input video clip $S=\{I_{S,i}\in\mathbb{R}^{3{\times}H{\times}W}|i=1,...,T\}$ consisting of $T$ frames. The key ideas are two-fold: 
On one hand, noticing that the high-level task of action recognition (\textit{e.g.,} pour milk) is defined as a combination of two low-level tasks, namely the hand motion (\textit{e.g.,} the movement of pouring) and the object in manipulation (\textit{e.g.,} the milk bottle)~\cite{garcia2018first,kwon2021h2o}, we follow this semantic hierarchy and divide $\vb{HTT}$ into two cascaded parts, namely the pose block $\vb{P}$ and the action block $\vb{A}$.
The pose block $\vb{P}$ first estimates the per-frame 3D hand pose and the interacting object category, and subsequently the action block $\vb{A}$ aggregates the predicted hand motion and object label over $S$ for action recognition. 

On the other hand, to cope with the different temporal granularities associated with the long-term action and the instantaneous pose, although we adopt transformer architecture for both $\vb{P}$ and $\vb{A}$, we focus $\vb{P}$ on a narrower temporal receptive field with only $t$ consecutive frames ($t<T$), while applying $\vb{A}$ over all the $T$ frames. 

\subsection{Hand Pose Estimation with Short-Term Temporal Cue}\label{sec:pose}

To overcome the frequent occlusion and truncation of hands in interaction in the egocentric view, we leverage temporal cues to improve the robustness of hand pose estimation.
Hand poses represent instant motions in time, so referring to a long time span can overemphasize the temporally distant frames, which could sacrifice the accuracy of local motion (see Sec.~\ref{sec:ablation} and the supplementary video for an ablation).
Therefore, we localize the time span for pose estimation by dividing the video clip $S$ into $m$ consecutive segments $\mathtt{seg_t}(S)=(\bar{S}_1,\bar{S}_2,...,\bar{S}_m)$, where $m=\lceil T/t \rceil$, $\bar{S}_i=\{I_{\bar{S}_i,j}=I_{S,k}\in{S} | k=(i-1)t+j, j=1,...,t\}$ (see Fig.~\ref{fig:network}); tokens beyond the length $T$ are padded but masked out from self-attention computation. 
This scheme can be regarded as a shifting window strategy with window size $t$.
The module $\vb{P}$ processes each segment $\bar{S}\in\mathtt{seg_t}(S)$ in parallel to capture the temporal cue for hand pose estimation.

For each local segment $\bar{S}\in\mathtt{seg_t}(S)$, $\vb{P}$ takes the sequence of per-frame ResNet features $(f(I_{\bar{S},1}),...,f(I_{\bar{S},t}))$ as input, and outputs a sequence $(g_{\bar{S}}(I_{\bar{S},1}),...,g_{\bar{S}}(I_{\bar{S},t}))$. 
The $j$-th token $g_{\bar{S}}(I_{\bar{S},j})\in\mathbb{R}^d$ ($j=1,...,t$) 
corresponds to the frame $I_{\bar{S},j}$ and encodes also the temporal cue from $\bar{S}$. 
We then decode the hand pose for $I\in\bar{S}$ from its temporal-dependent feature $g_{\bar{S}}(I)$:
\begin{equation}
P_I=(P^{2D}_I,P^{dep}_I)=\mathtt{MLP_1}(g_{\bar{S}}(I)))
\end{equation}
where for hand (or two-hands) with $J$ hand joints, $P$ is the concatenation of the joint coordinates in the 2D image plane $P^{2D}_I\in\mathbb{R}^{J\times{2}}$ and the joint depth to the camera $P^{dep}_I\in\mathbb{R}^{J\times{1}}$, while $\mathtt{MLP_1}$ has three layers of width $[d,d,3J]$ with LeakyReLu as the activation functions for the hidden layers. 
For supervision we compare the prediction with the groundtruth hand pose $(P^{2D}_{I,gt},P^{dep}_{I,gt})$ and minimize the $L1$-loss:
\begin{equation}\label{eq:hand}
    L_{H}(I)=\frac{1}{J}(||P^{2D}_I-P^{2D}_{I,gt}||_1 + \lambda_1 ||P^{dep}_I-P^{dep}_{I,gt}||_1)
\end{equation}
where $\lambda_1$ is a hyper-parameter to balance the different magnitudes of the 2D loss and the depth loss. The 3D positions of the hand joints in the camera space $P^{3D}_I\in\mathbb{R}^{J\times{3}}$ for $I$ can then be recovered given the camera intrinsics. 

As the category of the manipulated object supplies the noun of an action~\cite{tekin2019h+}, we also predict from $g_{\bar{S}}(I)$ the object category. 
Denoting the number of object classes as $n_o$, we obtain an $n_o$-dim classification probability vector from $g_{\bar{S}}(I)$ with another $\mathtt{MLP_2}$ with two layers of width $[d,n_o]$:
\begin{equation}
O_I=[p(o_1|I),...,p(o_{n_o}|I)]=\mathtt{softmax}(\mathtt{MLP_2}(g_{\bar{S}}(I))).
\end{equation}
Given the groundtruth object label $o_{gt}$, the target probability is defined as a one-hot vector $w^o(I)=(w^o_{I,1},...,w^o_{I,n_o})$ with only $w^o_{I,o_{gt}}=1$ for $o_{gt}$. 
The object classification task is supervised to minimize the cross entropy between the predicted $O_I$ and the groundtruth $w^o(I)$:
\begin{equation}\label{eq:olabel}
L_O(I)=-\sum_{i=1}^{n_o} w^o_{I,i} \log{p(o_i|I)}.
\end{equation}

\subsection{Action Recognition with Long-Term Temporal Cue}\label{sec:action}

Perceiving a long time span clarifies the action performed (Fig.~\ref{fig:teasor}). Therefore we have the action module $\vb{A}$ of $\vb{HTT}$ that leverages the full input sequence $S$ to predict the action label.

Specifically, the input of $\vb{A}$ is a sequence of $T+1$ tokens $(\vb{\alpha_{in}}, h(I_{S,1}), ..., h(I_{S,T}))$. 
We follow previous works~\cite{devlin2018bert,dosovitskiy2020image} to introduce a trainable token $\vb{\alpha_{in}}\in\mathbb{R}^d$ that aggregates the global information across $S$ for action classification. 
The remaining tokens encode the per-frame information of the hand pose and object label.
For a frame $I\in\bar{S}$ with $\bar{S}\in\mathtt{seg_t}(S)$, $h(I)\in\mathbb{R}^d$ mixes its 2D hand pose $P^{2D}_I$, the probability distribution of object classification $O_I$ and the image feature $g_{\bar{S}}(I)$ computed by $\vb{P}$, \textit{i.e.}:
\begin{equation}\label{eq:action_intoken}
    h(I)= \mathtt{FC_1}[\mathtt{FC_2}(P^{2D}_I),\mathtt{FC_3}(O_I), g_{\bar{S}}(I)]
\end{equation}
where $\mathtt{FC_2}(.)$ and $\mathtt{FC_3}(.)$ respectively output $d$-dim features, and $\mathtt{FC_1}[.,.,.]$ reduces the concatenation of three input features into $d$-dim to fit in the token dimension of $\vb{A}$.
Hence, $\vb{A}$ is expected to recognize the performed action based on the hand motion, the label of the manipulated object, and also the per-frame image features which may encode other useful cues like object appearance and hand-object contacts. We provide an ablation for the input feature leveraged by $\vb{A}$ in Sec.~\ref{sec:ablation} and Tab.~\ref{tab:ablation_action}.

To classify the action for $S$, we make use of the first token $\alpha_{out}\in\mathbb{R}^d$ of the output sequence by $\vb{A}$, where with an $\mathtt{FC_4}$ we regress the probability distribution over the given action taxonomy with $n_a$ pre-defined actions as:
\begin{equation}
\vb{A}(S)=[p(a_1|S),...,p(a_{n_a}|S)]=\mathtt{softmax}(\mathtt{FC_4}(\alpha_{out})).
\end{equation}
Given the groundtruth action $a_{gt}$ and the target probability as a one-hot vector $w(S)=(w_{S,1},...,w_{S,n_a})$ with only $w_{S,a_{gt}}=1$ for $a_{gt}$, 
we supervise by minimizing the cross entropy for classifying the action category as:
\begin{equation}
L_A(S)=-\sum_{i=1}^{n_a} w_{S,i} \log{p(a_i|S)}.
\end{equation}

To summarize, given an input video clip $S$ to $\vb{HTT}$, the total training loss is:
\begin{equation}\label{eq:action}
L={L_A}(S)+\frac{1}{T}\sum_{\bar{S}\in\mathtt{seg_t}(S)}\sum_{I\in\bar{S}}(\lambda_2{L_H}(I)+\lambda_3{L_O}(I))
\end{equation}
where $\lambda_2,\lambda_3$ are hyperparameters to balance different loss terms.

\subsection{Implementation Details}\label{sec:implementation}

\noindent\textbf{Network parameters } We set $T=128$ and $t=16$ as the maximum input sequence length for $\vb{A}$ and $\vb{P}$, respectively. All input images are resized to $H=270, W=480$. The ResNet feature, as well as the tokens of $\vb{P}$ and $\vb{A}$ are all of the dimension $d=512$. Both $\vb{P}$ and $\vb{A}$ have two transformer encoder layers, use the fixed sine/cosine position encoding~\cite{vaswani2017attention} and put layer normalization before the attention and feed-forward operations; each layer has 8 attention heads, and the feed-forward block has a dimension of 2048. 

The value of $T$ derives from the limitation of available computational resources, as simultaneously processing all frames for an input video $V$ with more than $T$ frames causes an out-of-memory error. 
To process a longer video, we split $V$ into a clip set $\mathtt{seg_T}(V)$ where each clip $S\in\mathtt{seg_T}(V)$ can be processed by $\vb{HTT}$. 
The construction of $\mathtt{seg_T}(V)$ has two steps: we first downsample $V$ into two sub-sequences with a sampling ratio of 2, and then divide both of the two sub-sequences into consecutive clips by adopting the shifting window strategy with a window size $T$.

\noindent\textbf{Training stage details} 
We set $\lambda_1=200,\lambda_2=0.5,\lambda_3=1$ to balance different loss terms, and train with the Adam optimizer~\cite{kingma2014adam} with an initial learning rate of $3\times{10}^{-5}$, where we halve the learning rate every 15 epochs. Same as~\cite{hasson2020leveraging}, the ResNet-18 is initialized with weights pretrained on ImageNet, which are updated during backpropagation, except that the batch normalization layers are frozen.
The whole network was trained on 2 GPUs with a total batch size of 2, and we observe convergence after 45 epochs. We also follow~\cite{hasson2020leveraging} to add online augmentation to input image sequences, by randomly adjusting the hue, saturation, contrast, and brightness, as well as adding random Gaussian blur and 2D translation.

\begin{figure}[!tb]
\centering
\includegraphics[width=0.95\linewidth]{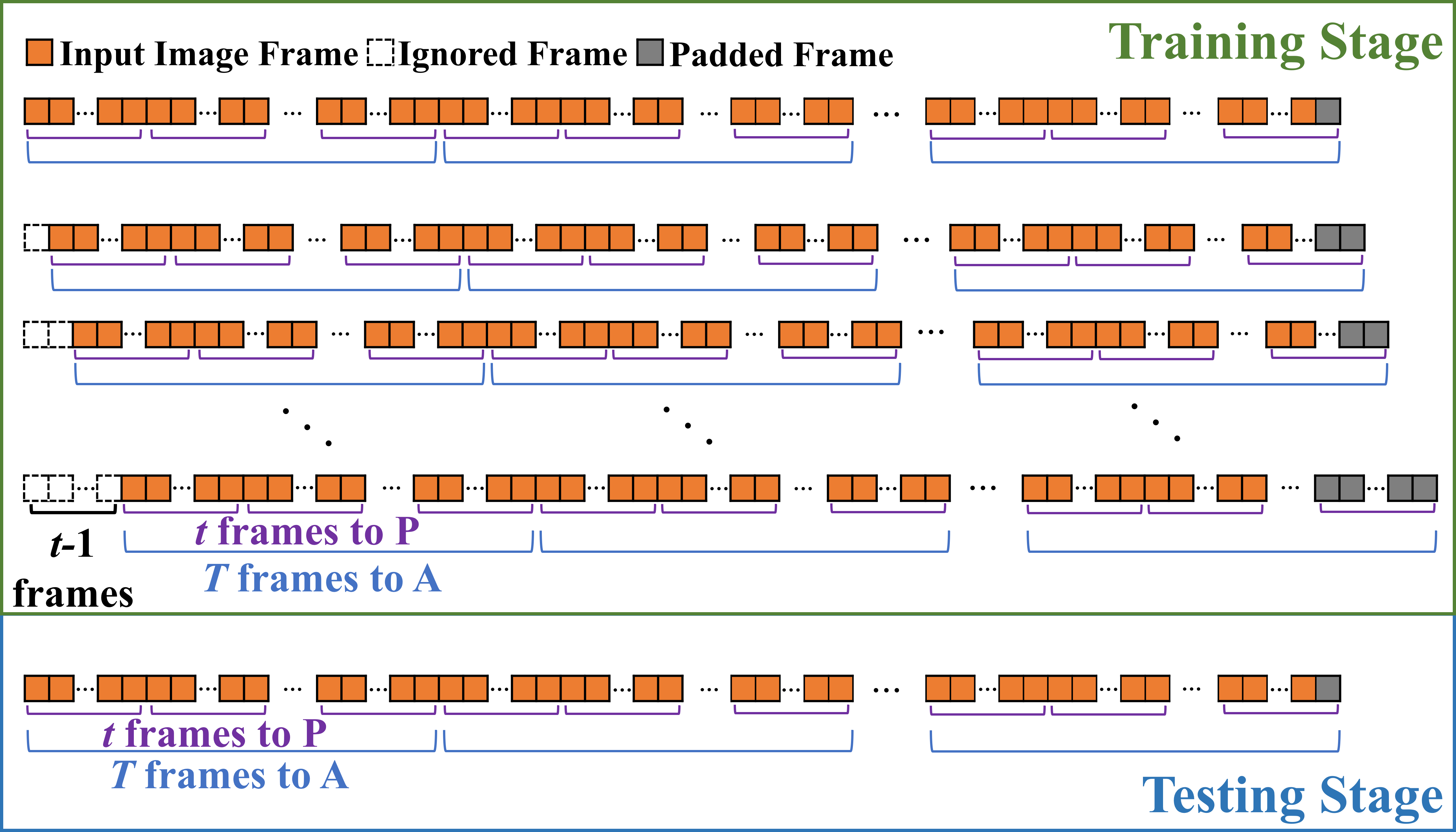}
\vspace{-2mm}
\caption{Segmentation strategy for dividing a long video into inputs of $\vb{HTT}$.
In the testing stage, we start from the first frame, while in the training stage, we offset the starting frame within $t$ frames to augment the training data diversity.}\label{fig:sw}
\vspace{-2mm}
\end{figure}

To augment sampling variations of training data, we offset the starting frame to each of the first $t$ frames, as illustrated in Fig.~\ref{fig:sw}.
Offsets within the $t$ segment ensure that both $\vb{P}$ and $\vb{A}$ consume different augmented data generated from the same sequence; in contrast, an offset larger than $t$ will cause duplicated segments fed to $\vb{P}$.

\noindent\textbf{Testing stage computation } 
For a video with more than $T$ frames, we first obtain the set $\mathtt{seg_T}(V)$ for the original test video $V$, then feed each $S\in\mathtt{seg_T}(V)$ to $\vb{HTT}$, and obtain the per-image 3D hand pose from the output of $\vb{P}$ and the action category for $V$ by voting from the output category among $S\in\mathtt{seg_T}(V)$ (Fig.~\ref{fig:sw}). 
Note that in this way we achieve efficient computation as each image is processed only once by both $\vb{P}$ and $\vb{A}$.

\section{Experiments}

\subsection{Dataset}
We train and test the proposed method on two public datasets FPHA~\cite{garcia2018first} and H2O~\cite{kwon2021h2o} for 3D hand pose estimation and action recognition from first-person views. Both datasets are recorded in multiple indoor scenarios and have a frame rate of 30 fps. We use the groundtruth labels for hand pose, action, and object category provided by the datasets for supervision and evaluation. 

\noindent\textbf{FPHA~\cite{garcia2018first}} This dataset records 6 subjects performing 45 actions, interacting with 26 rigid or non-rigid objects. It annotates $J=21$ joints for only the subject's right hand, with the hand pose data collected from the wearable magnetic sensors. Following the \textit{action split}~\cite{garcia2018first,tekin2019h+,yang2020collaborative} we train our method on 600 videos and test it on 575 videos. All subjects and actions are seen at both the training and testing stages.

\noindent\textbf{H2O\cite{kwon2021h2o}} This dataset has 4 subjects performing 36 actions related to manipulating 8 objects. Markerless 3D annotations for both hands with a total number of $J=21\times{2}$ joints are provided. We follow~\cite{kwon2021h2o} to use egocentric view sequences with annotated actions for training and testing, where the training split has 569 videos including all actions for the first 3 subjects, and the testing split has 242 videos of the remaining subject unseen in training.

\subsection{Metrics}
\noindent\textbf{Action recognition} We report the \textit{Classification Accuracy} over the test split, by comparing the predicted and groundtruth action categories for each test video.

\noindent\textbf{3D hand pose estimation} We evaluate the proposed method by comparing its estimated 3D joint positions with the groundtruth. On FPHA~\cite{garcia2018first}, we follow the baseline methods~\cite{tekin2019h+,yang2020collaborative,fan2020adaptive} to report the \textit{Percentage of Correct Keypoints (PCK)} under different error thresholds for joints~\cite{zimmermann2017learning} and the corresponding \textit{Area Under the Curve (AUC)}. We respectively report \textit{3D PCK} and \textit{3D PCK-RA} for evaluation in the camera space and the root-aligned (RA) space where for each frame the estimated wrist is aligned with its groundtruth position. 

On H2O~\cite{kwon2021h2o}, we follow its benchmark~\cite{kwon2021h2o} to evaluate in the camera space, by reporting the \textit{3D PCK} metric and the \textit{Mean End-Point Error (MEPE)} for hands~\cite{zimmermann2017learning}. 
We also establish a baseline for the root-aligned space, where for each frame we align the estimated wrist with its groundtruth position and report the corresponding \textit{3D PCK-RA} and \textit{MEPE-RA} for the non-wrist joints.

\subsection{Comparison with Related Works}
\noindent\textbf{Comparison on FPHA~\cite{garcia2018first}} 
Our closest state-of-the-art baseline method is Collaborative~\cite{yang2020collaborative} which leverages the temporal cue for both 3D hand pose estimation and action recognition. 
Another baseline method is H+O~\cite{tekin2019h+}, which first conducts single image-based pose estimation and then exploits the temporal domain for action recognition.
Moreover, for hand pose estimation, we also compare our method with ACE-Net~\cite{fan2020adaptive} which focuses only on video-based 3D hand pose estimation and reports on FPHA~\cite{garcia2018first}. 
Among RGB-based methods that tackle only the action recognition, we add baseline methods \cite{hu2015jointly,feichtenhofer2016convolutional} from FPHA~\cite{garcia2018first} to our discussion. 

\begin{figure}[!tb]
\centering
\includegraphics[width=0.53\linewidth]{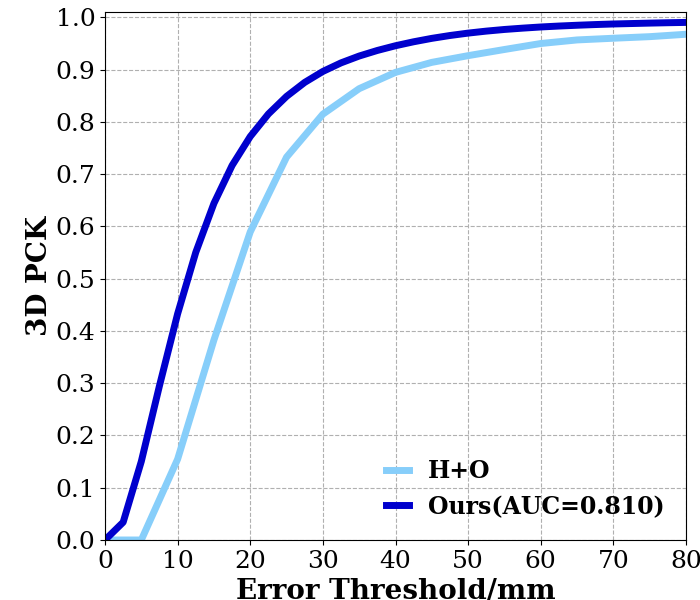}
\hfill
\includegraphics[width=0.45\linewidth]{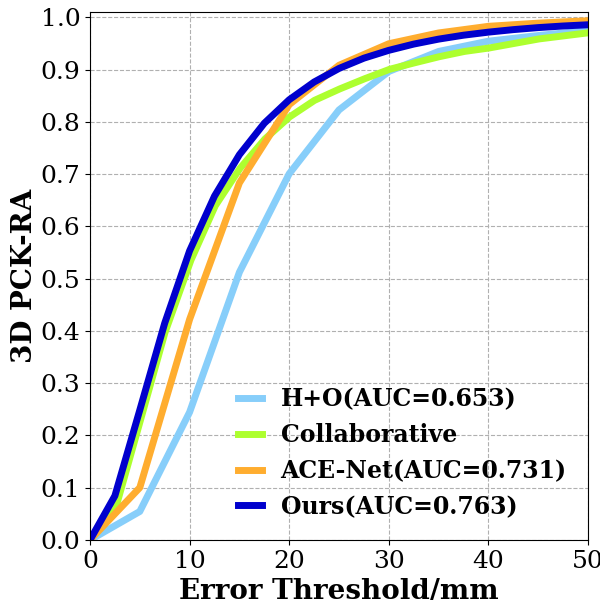}
\caption{3D PCK(-RA) of hand pose estimation on FPHA~\cite{garcia2018first}. We report the 3D PCK(-RA) versus different error thresholds by respectively evaluating in the camera space (Left figure) and the root-aligned space (Right figure). 
}\label{fig:fpha_hand}
\end{figure}

\begin{figure*}[!tb]
\vspace{-2mm}
\centering
\includegraphics[width=0.95\linewidth]{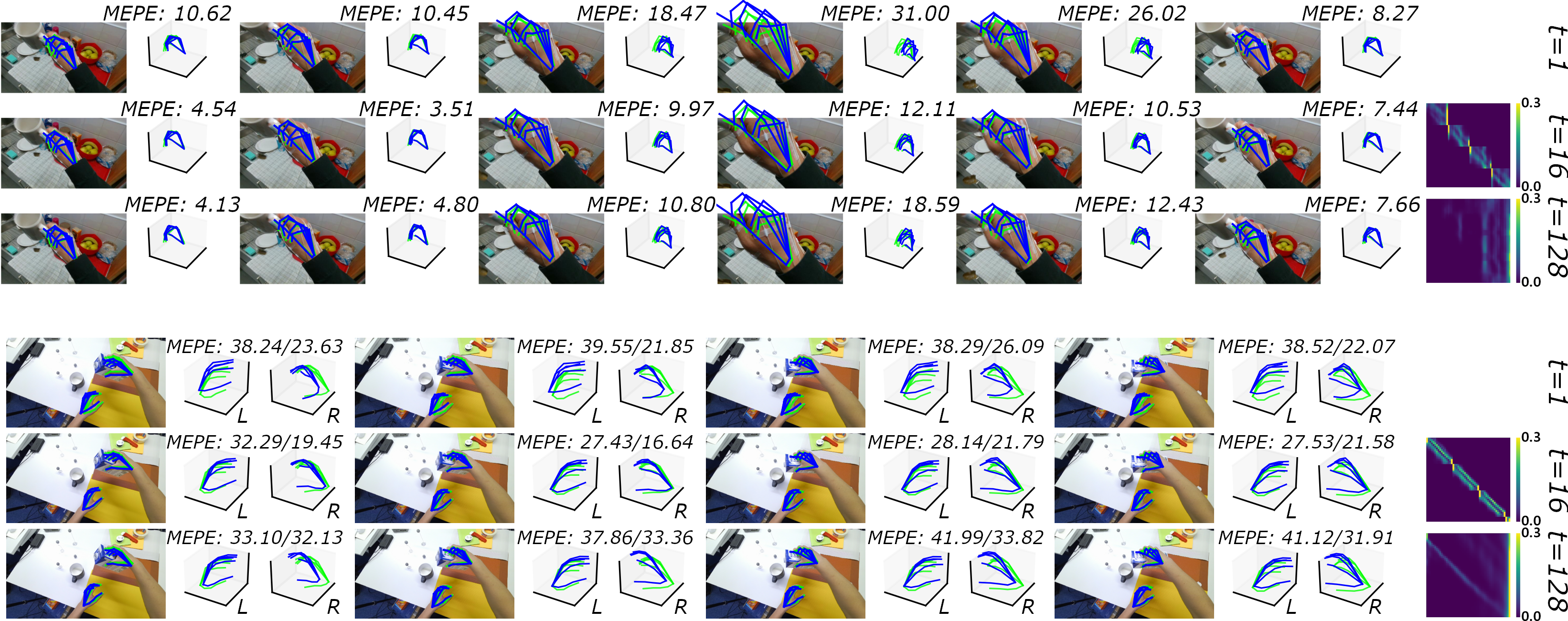}
\vspace{-2mm}
\caption{Qualitative cases of 3D hand pose estimation in the camera space and its 2D projection for sequential images from FPHA~\cite{garcia2018first}(\textit{Upper}) and H2O~\cite{kwon2021h2o}(\textit{Lower}). We show our choice of leveraging the short-term temporal cue with $t=16$, and compare it with the image-based baseline of $t=1$, and the setting using a long-term temporal cue with $t=128$. The estimation and GT are respectively denoted in blue and green, where the MEPE with unit $mm$ is also attached. 
For $t=16,128$, the attention weights in the final layer of $\vb{P}$ are visualized. 
Our $t=16$ shows enhanced robustness under invisible joints compared with $t=1$, while avoiding over-attending to distant frames and ensuring sharp local motion compared with a long-term $t=128$.}\label{fig:quali}
\end{figure*}

We demonstrate the effectiveness of our method with competitive results for both action recognition and hand pose estimation, as shown in Tab.~\ref{tab:fpha_action} and Fig.~\ref{fig:fpha_hand}. 
For accuracy of action recognition, we outperform baseline methods by respectively improving for more than 8\% over Collaborative~\cite{yang2020collaborative} and 11\% over H+O~\cite{tekin2019h+}. 
For hand pose estimation, Fig.~\ref{fig:fpha_hand} shows the 3D PCK and the 3D PCK-RA, where we note that only H+O~\cite{tekin2019h+} reports results in the camera space. 
Compared with our closest related work Collaborative~\cite{yang2020collaborative}, we improve the performance with a notable margin for error thresholds $>15mm$. We also show better accuracy than H+O~\cite{tekin2019h+} at all reported error thresholds, and outperform ACE-Net~\cite{fan2020adaptive} for error thresholds less than $20mm$ and the AUC metrics.
Unfortunately, the comparing methods have not yet released their code or data for inspecting the specific cases, but from our qualitative cases for 3D pose estimation in Fig.~\ref{fig:quali} and the supplementary video we can see that our results are robust to self-occlusions and truncations that are common in egocentric view (more discussions in Sec.~\ref{sec:ablation}), which can be the primary reason for the differences.

\begin{table}[!tb]
\begin{center}
\resizebox{0.99\linewidth}{!}{
\begin{tabular}{|c|c|c|c|c|c|}
    \hline
    &\tabincell{c}{Joule-color\\~\cite{hu2015jointly}} & \tabincell{c}{Two-stream\\~\cite{feichtenhofer2016convolutional}} & \tabincell{c}{H+O\\~\cite{tekin2019h+}}& \tabincell{c}{Collaborative\\~\cite{yang2020collaborative}} & Ours \\
    \hline
    Accuracy & 66.78 & 75.30 &  82.43 & 85.22 & \textbf{94.09} \\
    \hline
\end{tabular}}
\caption{Classification accuracy of action recognition for RGB-based methods on FPHA~\cite{garcia2018first}. }\label{tab:fpha_action}
\end{center}
\vspace{-6mm}
\end{table}

\noindent\textbf{Comparison on H2O~\cite{kwon2021h2o}} We focus on comparison with H+O~\cite{tekin2019h+} and H2O~\cite{kwon2021h2o}, as these two methods output both the per-frame 3D hand pose and the action category for the input video. 
For each task, we additionally compare our method with other RGB-based methods that solve only the given task: for action recognition, C2D~\cite{wang2018non}, I3D~\cite{carreira2017quo} and SlowFast~\cite{feichtenhofer2019slowfast} are compared; for hand pose estimation, we add LPC~\cite{hasson2020leveraging} into the comparison.
Results of these baseline methods are copied from the benchmark of H2O~\cite{kwon2021h2o}. 
We note that LPC~\cite{hasson2020leveraging}, H+O~\cite{tekin2019h+} and H2O~\cite{kwon2021h2o} conduct image-based pose estimation. Furthermore, H+O~\cite{tekin2019h+} and LPC~\cite{hasson2020leveraging} train separate networks for each hand, while H2O~\cite{kwon2021h2o} and ours estimate for both hands within a unified network.

\begin{figure}[!tb]
\centering
\includegraphics[width=0.53\linewidth]{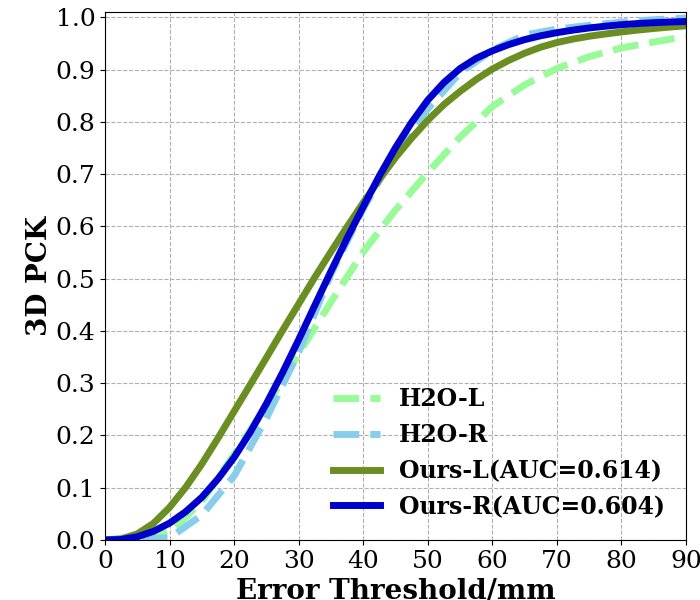}
\hfill
\includegraphics[width=0.45\linewidth]{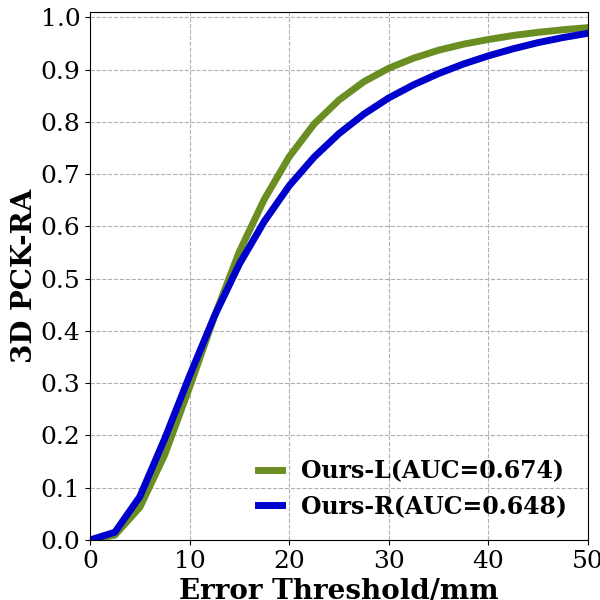}
\caption{3D PCK(-RA) of hand pose estimation on the test split of H2O~\cite{kwon2021h2o}. We report the 3D PCK(-RA) versus different error thresholds by respectively evaluating in the camera space (Left figure) and the root-aligned space (Right figure).}\label{fig:h2o_pck}
\end{figure}

We first report the accuracy of action recognition in Tab.~\ref{tab:h2o_action}, where we outperform baseline methods and improve the state-of-the-art setup proposed in H2O~\cite{kwon2021h2o} with a margin over 7\%. 
For 3D hand pose estimation, we first follow the H2O benchmark~\cite{kwon2021h2o} to evaluate the results in the camera space, reporting the 3D PCK and MEPE in Fig.~\ref{fig:h2o_pck} and Tab.~\ref{tab:h2o_epe}, respectively. 
Our better performance demonstrates the importance and benefits of leveraging the temporal coherence for robust hand pose estimation under frequent occlusions in the egocentric view.
We further establish a baseline by evaluating the results in the root-aligned space, and report our MEPE-RA and 3D PCK-RA in Tab.~\ref{tab:h2o_epe} and Fig.~\ref{fig:h2o_pck}, respectively.
We show our qualitative cases in Fig.~\ref{fig:quali} and the supplementary video.

\begin{table}[!tb]
\begin{center}
\resizebox{0.99\linewidth}{!}{
\begin{tabular}{|c|c|c|c|c|c|}
    \hline
    &\multicolumn{4}{c|}{MEPE in Camera Space} & MEPE-RA \\
    \hline
    & H+O~\cite{tekin2019h+}&LPC~\cite{hasson2020leveraging}& H2O~\cite{kwon2021h2o} & Ours & Ours \\
    \hline
    Left & 41.42 & 39.56 & 41.45 & \textbf{35.02} & 16.59 \\
    \hline
    Right & 38.86& 41.87 & 37.21 & \textbf{35.63} & 17.91\\
    \hline
\end{tabular}}
\caption{MEPE and MEPE-RA of hand pose estimation on the test split of H2O~\cite{kwon2021h2o}, the unit is $mm$.}\label{tab:h2o_epe}
\end{center}
\end{table}

\begin{table*}[!tb]
\begin{center}
\resizebox{0.9\linewidth}{!}{
\begin{tabular}{|c|c|c|c|c|c|c|c|}
    \hline
    & C2D~\cite{wang2018non} & I3D~\cite{carreira2017quo} & SlowFast~\cite{feichtenhofer2019slowfast}& H+O~\cite{tekin2019h+}& \tabincell{c}{H2O w/ ST-GCN~\cite{yan2018spatial}} & \tabincell{c}{H2O w/ TA-GCN~\cite{kwon2021h2o} } & Ours \\ 
    \hline
    Accuracy & 70.66 & 75.21 & 77.69 & 68.88 & 73.86 &79.25 & \textbf{86.36} \\
    \hline
\end{tabular}}
\caption{Classification accuracy of action recognition for RGB-based methods on H2O~\cite{kwon2021h2o}.}\label{tab:h2o_action}
\end{center}
\end{table*}

\subsection{Ablation Study}~\label{sec:ablation}
We verify our key design choices with ablation study on both the FPHA~\cite{garcia2018first} and H2O~\cite{kwon2021h2o} datasets. For pose estimation, we report MEPE and AUC for 3D PCK versus different error thresholds, with regard to both the camera space and per-frame root-aligned space. For action, we report the classification accuracy.

\noindent\textbf{Short-term temporal cue for pose } We first demonstrate the benefits brought by leveraging temporal cue for pose estimation, where we remove the pose block $\vb{P}$ and conduct image-based pose estimation by regressing the hand pose and object label from the ResNet feature. Thus, we have the temporal window size $t=1$ for pose and keep $T=128$ for action. We report the evaluation of the 3D hand pose for this setup in Tab.~\ref{tab:ablation_pose}, whose degraded performance in both camera space and root-aligned space for both datasets validates the benefits brought by exploiting short-term temporal cues with $t>1$. 
Moreover, we qualitatively compare the two setups with $t=1$ and our $t=16$ in Fig.~\ref{fig:quali} and the supplementary video, where with temporal cue we see improved robustness under occlusion and truncation.

We then move on to discuss the impact of various time spans $t$, and report in Tab.~\ref{tab:ablation_pose} for hand pose estimation. In the camera space, $t=16$ shows the best performance while longer time spans degrade accuracy, which reveals that attending to temporally distant frames in a long time span can override the local sharpness for motion. 
For further validation, we present qualitative cases in Fig.~\ref{fig:quali} and the supplementary video, where we compare the two setups of $t=T=128$ and our $t=16$ by visualizing the hand pose estimation and the attention map for the final layer of $\vb{P}$, which confirms that the long time-spans cause diffused attention maps that cannot capture localized motions. 
Moreover, in the per-frame root-aligned space, leveraging a local time span with $t$ as 32 or 64 report the best results, while our choice of using a smaller $t=16$ shows comparable performance. 
These results validate our choice of $t=16$ for robust and accurate pose estimation.
 
\begin{table*}[!tb]
\begin{center}
\resizebox{0.85\linewidth}{!}{
\begin{tabu}{|c|c|c|c|c|c|c|c|c|c|c|c|c|}
\hline
&\multicolumn{4}{c|}{FPHA~\cite{garcia2018first}} & \multicolumn{8}{c|}{H2O~\cite{kwon2021h2o}} \\
\hline
\multirow{3}{*}{$t$} & \multicolumn{2}{c|}{In Camera Space} & \multicolumn{2}{c|}{In Root-Aligned Space}  & \multicolumn{4}{c|}{In Camera Space} & \multicolumn{4}{c|}{In Root-Aligned Space}\\
\cline{2-13}
& \multirow{2}{*}{AUC(0-80)} & \multirow{2}{*}{MEPE} & \multirow{2}{*}{AUC-RA(0-50)} & \multirow{2}{*}{MEPE-RA} & \multicolumn{2}{c|}{AUC(0-90)} & \multicolumn{2}{c|}{MEPE} & \multicolumn{2}{c|}{AUC-RA(0-50)} & \multicolumn{2}{c|}{MEPE-RA} \\ 
\cline{6-13}
&&&&&Left&Right&Left&Right&Left&Right&Left&Right\\
\hline
1 & 0.776 & 18.78 & 0.707 & 15.01 & 0.563  & 0.551 & 40.12 & 40.62 & 0.648 & 0.590 & 17.97 & 21.22\\
\hline
8 & 0.802 & 16.54 & 0.750 & 12.77 & 0.607  & 0.600 & 35.91 & 36.00 & 0.674 & 0.629 & 16.58  & 18.89 \\
\hline
16 & \textbf{0.810} & \textbf{15.81} & 0.763   & 12.13 &\textbf{0.614}  & \textbf{0.604} & \textbf{35.02}&  \textbf{35.63} & 0.674  & 0.648 & 16.59 & 17.91\\
\hline
32 & 0.805 & 16.14 & 0.766 & 11.96 & 0.594 & 0.573 & 36.78 & 38.48 & 0.676 & \textbf{0.650} & 16.56 & \textbf{17.85} \\
\hline
64 & 0.805 & 16.19 & \textbf{0.767} & \textbf{11.92} & 0.599 & 0.567 & 36.47 & 39.02 & \textbf{0.684} & 0.643 & \textbf{16.23} & 18.10\\
\hline
128 & 0.800 & 16.73 & 0.761 & 12.20 & 0.599 & 0.564 & 36.41 & 39.36 & 0.673 & 0.636 & 16.71 & 18.66 \\
\hline
\end{tabu}}
\caption{Ablation on time span $t$ for pose block $\vb{P}$ on the FPHA~\cite{garcia2018first} and H2O~\cite{kwon2021h2o} dataset. Reported are the AUC(0-$E$) with $E\in\{50,80,90\}$ for 3D PCK(-RA) at error thresholds ranging from 0 to $E$ $mm$ and the MEPE(-RA) in the unit of $mm$. All setups have $T=128$ for the action block $\vb{A}$.}\label{tab:ablation_pose}
\end{center}
\end{table*}

\begin{figure*}[!tb]
\vspace{-2mm}
\centering
\includegraphics[width=0.9\linewidth]{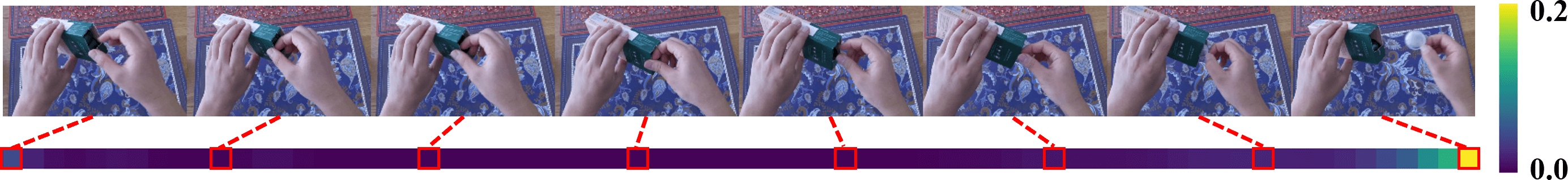}
\vspace{-2mm}
\caption{Visualization for weights of attention in the final layer of $\vb{A}$, from the action token to the frames. Presented is a video of \textit{take out espresso}, whose down-sampled image sequence is shown in the top row. The last few frames are the key to recognizing the action; in response, our network pays the most attention to these frames. The full image sequence of this demo can be found in the supplementary video.}\label{fig:attn_action}
\vspace{-2mm}
\end{figure*}

\noindent\textbf{Long-term temporal cue for action } Given the necessity of temporal information for resolving action ambiguity (see Fig.~\ref{fig:teasor}), we first verify the benefits of using a long time span for action, by varying $T$ from $T=t=16$ to $T=128$. 
We report the accuracy for both datasets in the upper part of Tab.~\ref{tab:ablation_action}. Unlike per-frame hand pose estimation which benefits from a short-term temporal cue, here longer time spans improve the accuracy of action recognition, which supports our design of leveraging different time spans for tasks with different temporal granularity.

For our $T=128$, we also visualize in Fig.~\ref{fig:attn_action} the attention weights from the action token to the per-frame tokens in the final layer of $\vb{A}$, obtained on a video for \textit{take out espresso} whose action can only be judged by the last few frames depicting the process of taking the capsule out of the box. We observe that these keyframes for action recognition are given the most attention. More cases for various actions are provided in the supplementary video, where the distribution of attention weights shows clear correspondence patterns with respect to different actions.

\noindent\textbf{Cascaded hierarchy for action recognition } To echo the ``\textit{verb+noun}" pattern described by action labels, we cascade the two blocks $\vb{P}$ and $\vb{A}$ to classify the action category, and feed the per-frame hand pose, object label, and image feature obtained by $\vb{P}$ as input for $\vb{A}$ (see Eq.~\ref{eq:action_intoken}).
We first verify the necessity of the cascaded design. For comparison, we set $\vb{P}$ and $\vb{A}$ in a parallel structure, where the ResNet image feature is the per-frame input token for both $\vb{P}$ and $\vb{A}$. We observe degraded performance with this parallel design, as reported in the middle part of Tab.~\ref{tab:ablation_action}.

Then based on our cascaded design, we ablate the components of the per-frame input to $\vb{A}$. As reported in the lower part in Tab.~\ref{tab:ablation_action}, our full version shows the best performance, which demonstrates the benefits of modeling the semantic correlation among hand pose, object label, and action, as well as the gains of letting $\vb{A}$ exploit also the other miscellaneous cues encoded in the image feature.

\begin{table}[!tb]
\begin{center}
\resizebox{0.99\linewidth}{!}{
\begin{tabu}{|c|c|c|c|c|c|c|}
\hline
\multirow{2}{*}{$T$} & \multirow{2}{*}{Cascaded $\vb{P},\vb{A}$} &
\multicolumn{3}{c|}{Input Feature for $\vb{A}$} &
\multicolumn{2}{c|}{Classification Accuracy} \\
\cline{3-7}
 & & Image Feature & Hand Pose & Object Label& FPHA~\cite{garcia2018first} &  H2O~\cite{kwon2021h2o}\\
\tabucline[1.2pt]{-}
16 & $\checkmark$ & $\checkmark$ & $\checkmark$ & $\checkmark$ & 90.96 & 74.38\\
\hline
32 & $\checkmark$ & $\checkmark$& $\checkmark$ & $\checkmark$ & 91.65 & 79.34 \\
\hline
64 & $\checkmark$ & $\checkmark$& $\checkmark$ & $\checkmark$  & 92.35 & 78.51\\
\hline
128 & $\checkmark$ & $\checkmark$ & $\checkmark$ & $\checkmark$ & \textbf{94.09} & \textbf{86.36}\\
\tabucline[1.2pt]{-}
128 & &  $\checkmark$ &  &  & 93.22 & 80.17 \\
\hline
128 & $\checkmark$ & $\checkmark$ & $\checkmark$ & $\checkmark$ & \textbf{94.09} & \textbf{86.36}\\
\tabucline[1.2pt]{-}
128 & $\checkmark$& $\checkmark$ & $\checkmark$ &  & 93.57 & 82.65 \\
\hline
128 & $\checkmark$& $\checkmark$ &  & $\checkmark$ & 91.65 & 85.12 \\
\hline
128 & $\checkmark$& & $\checkmark$ & $\checkmark$ & 90.26 & 75.21\\
\hline
128 & $\checkmark$ & $\checkmark$ & $\checkmark$ & $\checkmark$ & \textbf{94.09} & \textbf{86.36}\\
\hline
\end{tabu}}
\caption{Ablation on key designs for action recognition. Reported is the accuracy on the FPHA~\cite{garcia2018first} and H2O~\cite{kwon2021h2o} dataset. All setups have $t=16$ for the pose block $\vb{P}$.}\label{tab:ablation_action}
\end{center}
\vspace{-6mm}
\end{table}
\section{Conclusion}
In this paper, we have proposed a unified framework to simultaneously handle the tasks of 3D hand pose estimation and action recognition for an egocentric RGB video. Our core framework is a hierarchical temporal transformer that has two cascaded parts, where the first one refers to a relatively short time span to output per-frame 3D hand pose and object label with enhanced robustness under occlusion and truncation, and the second one exploits the long-term temporal cue by aggregating the per-frame information for action recognition. 
In this way, we model the correlation between the two tasks with different semantic levels, and leverage different time spans according to their temporal granularity, both contributing to improved performances as verified by ablation studies.
Extensive evaluations on two first-person hand action benchmarks demonstrate the effectiveness of our method.

\noindent\textbf{Limitation and future work} We mainly focus on learning to exploit the temporal dimension for pose estimation and action recognition with a transformer-based framework, but adaptively modeling the spatial interactions of hand joints and objects (\textit{e.g.,} via a transformer module rather than a ResNet feature extractor) may further improve the performance, which we leave as future work. 
Another interesting direction is that our hierarchical sequential framework can potentially be extended to model motion prediction and generation, which are fundamental to tasks such as early action detection and forecasting for human-robot collaboration.

%\addtolength{\textheight}{-12cm}   % This command serves to balance the column lengths on the last page of the document manually. It shortens the textheight of the last page by a suitable amount. This command does not take effect until the next page so it should come on the page before the last. Make sure that you do not shorten the textheight too much.
%\section*{APPENDIX}
\noindent\textbf{Acknowledgments} This work is partially funded by the Research Grant Council of
Hong Kong (GRF 17210222). This work is also partially supported by the Innovation and Technology Commission of the HKSAR Government under the InnoHK initiative.

%%%%%%%%% REFERENCES
%{\small
%\bibliographystyle{ieee_fullname}
%\bibliography{egbib}
%}

\end{document}